\documentclass[runningheads]{llncs}

\makeatletter
\newcommand{\@chapapp}{\relax}%
\makeatother

\usepackage[title]{appendix}
\usepackage{caption}
\usepackage{graphicx}

\newcommand{\ripper}{\texttt{Ripper}}

\begin{document}

\title{An Investigation into Mini-Batch Rule Learning}

\author{Florian Beck\inst{1} \and Johannes F\"urnkranz\inst{1}}

\authorrunning{F. Beck, J. F\"urnkranz}

\institute{Institute for Application-oriented Knowledge Processing (FAW) \\ JKU Linz, Austria\\
\email{\{fbeck,juffi\}@faw.jku.at}}

\maketitle

\begin{abstract}
We investigate whether it is possible to learn rule sets efficiently in a network structure with a single hidden layer using iterative refinements over mini-batches of examples. A first rudimentary version shows an acceptable performance on all but one dataset, even though it does not yet reach the performance levels of \ripper.

\keywords{Rule learning \and Mini-batch training}
\end{abstract}

\section{Introduction}
\ripper \cite{DBLP:conf/icml/Cohen95} was the first rule learning system that effectively countered the overfitting problem and is still a state-of-the-art method in inductive rule learning \cite{DBLP:series/cogtech/FurnkranzGL12}. It scales nearly linearly with the number of examples in a dataset and therefore outperforms other rule learning algorithms such as \texttt{C4.5rules} in terms of runtime. While linear complexity is uncritical for small datasets, in bigger datasets it might be insufficient. 

Neural networks cope with this problem and often even score a better accuracy, whereas at the same time they have the disadvantage of being difficult to interpret. There are already some approaches that combine the efficiency of neural networks with the interpretability of rules. Algorithms like ENDER \cite{DBLP:journals/datamine/DembczynskiKS10}, Neural Logic Programming \cite{DBLP:journals/corr/YangYC17}, or BOOMER \cite{mr:ECML-PKDD-2020} learn rule sets or lists by minimizing a loss by using gradient descent and differentiable models like in neural networks. Another approach \cite{DBLP:conf/icml/RuckertK03} applies stochastic local search to find good solutions in a reasonable period of time. With very fast decision rules (\texttt{VFDR}) \cite{DBLP:journals/datamine/KosinaG15} there already exists an algorithm especially adapted to data streams, which can also handle changes in the data over time. 

In this short communication, we focus on an evaluation of the mini-batch approach as one of the efficiency-improving techniques of neural networks. A network tailored to the learning of rule sets could use mini-batches to efficiently learn rule sets itself. Moreover, repeated fine-tuning on small datasets could replace the expensive optimization part of the \ripper algorithm. 
It could as well be expanded to even deeper rule structures with invention of new features in hidden layers.
Finally, this approach also brings the benefit of being an anytime algorithm.

\section{Network for Mini-Batch Rule Learning}
We created a network whose output should be structured like a rule set in \texttt{DNF}. Therefore, the network consists of three layers, the input layer, one hidden layer (= \texttt{AND} layer) and the output layer (= \texttt{OR} layer). The \texttt{AND} layer combines the literals to rules and the \texttt{OR} layer rules to a rule set. 
The network is designed for binary classification problems and produces a single prediction output that is either \texttt{True} or \texttt{False} whereby \texttt{True} means that the less frequent class is chosen. 

In the input layer, the network currently only takes nominal attributes $\mathcal{A}$ into account and converts them by one-hot-encoding to $|A_i|$ binary features (= literals) where $|A_i|$ is the number of possible values for attribute $A_i \in \mathcal{A}$. We also do this for Boolean attributes or other attributes with $|A_i|=2$ because it provides an option to encode unknown values by setting all corresponding features to \texttt{False}.
For the \texttt{AND} and \texttt{OR} layers, Boolean matrix multiplication is used to propagate the data through the network. Since matrix multiplication computes the sum of component-wise products, it corresponds to a disjunction. To get the conjunction, according to De Morgan's law, the values can be inverted before and after the addition. 

Before the network is trained, the \texttt{AND} layer is initialized by adding literals randomly to the rules, i.e., by setting the corresponding layer variable to \texttt{True}. Each attribute is added with the probability $3/|\mathcal{A}|$ so that on average three literals of different attributes are added. The \texttt{OR} layer is initialized by \texttt{True}s, which means all rules belong to the rule set. 

During the training the data is processed in mini-batches, where in each iteration, greedy adjustments can be made to the current rule set. For each mini-batch a maximum of two flips is established to prevent overfitting. A flip is understood to be a change in the \texttt{AND} layer from \texttt{True} to \texttt{False} (= removal of a literal) or from \texttt{False} to \texttt{True} (= addition of a literal). In this case, other literals of the same attribute are removed automatically to avoid the conjunction of contradictory literals. The flip with the biggest improvement of the accuracy on this mini-batch is selected (if any improvement is possible). 

The last selection of the training is the re-selection of rules based on not only a single mini-batch, but on all training data. The \texttt{OR} layer is filled with \texttt{False}s and then the computed rules are greedily added as before the literals in the \texttt{AND} layer. Therefore, the rule set only contains rules that improve the accuracy on the total training data.

\section{Experiments}
For our first experiments, we selected the \textsf{adult} dataset from the UCI Repository with almost 50,000 instances of which 20\% were retained as a test set. The number of rules was set to 20 and the mini-batch size to 400 which resulted in nearly 100 mini-batches. The development of the accuracy across the mini-batches is shown in Fig.~\ref{fig1}. The accuracy on the current mini-batch is shown in blue, the accuracy on the training set in orange and the accuracy on the test set in green. After a sharp increase in the first mini-batch, the accuracies on the training and test set continue to rise until after 25 mini-batches their maximum values are almost reached. The training set accuracy runs slightly above the test set whereas the evaluation values on the mini-batches are furthest up since the rule set was adapted on this mini-batch immediately before.

Furthermore, the rule network and \ripper were applied and compared on ten datasets from Kaggle challenges and the UCI repository (Table~\ref{tab1}). For \ripper, the \texttt{Java} implementation \texttt{JRip} is used whereas the rule network is implemented in Python. The number of rules was again set to 20, the batch size to the square root of the number of instances. The accuracies were obtained via 1$\times$10-fold cross validation.

\vspace{-0.9em}
\begin{table}
\begin{minipage}{0.58\linewidth}
  \centering
  \includegraphics[width=\textwidth]{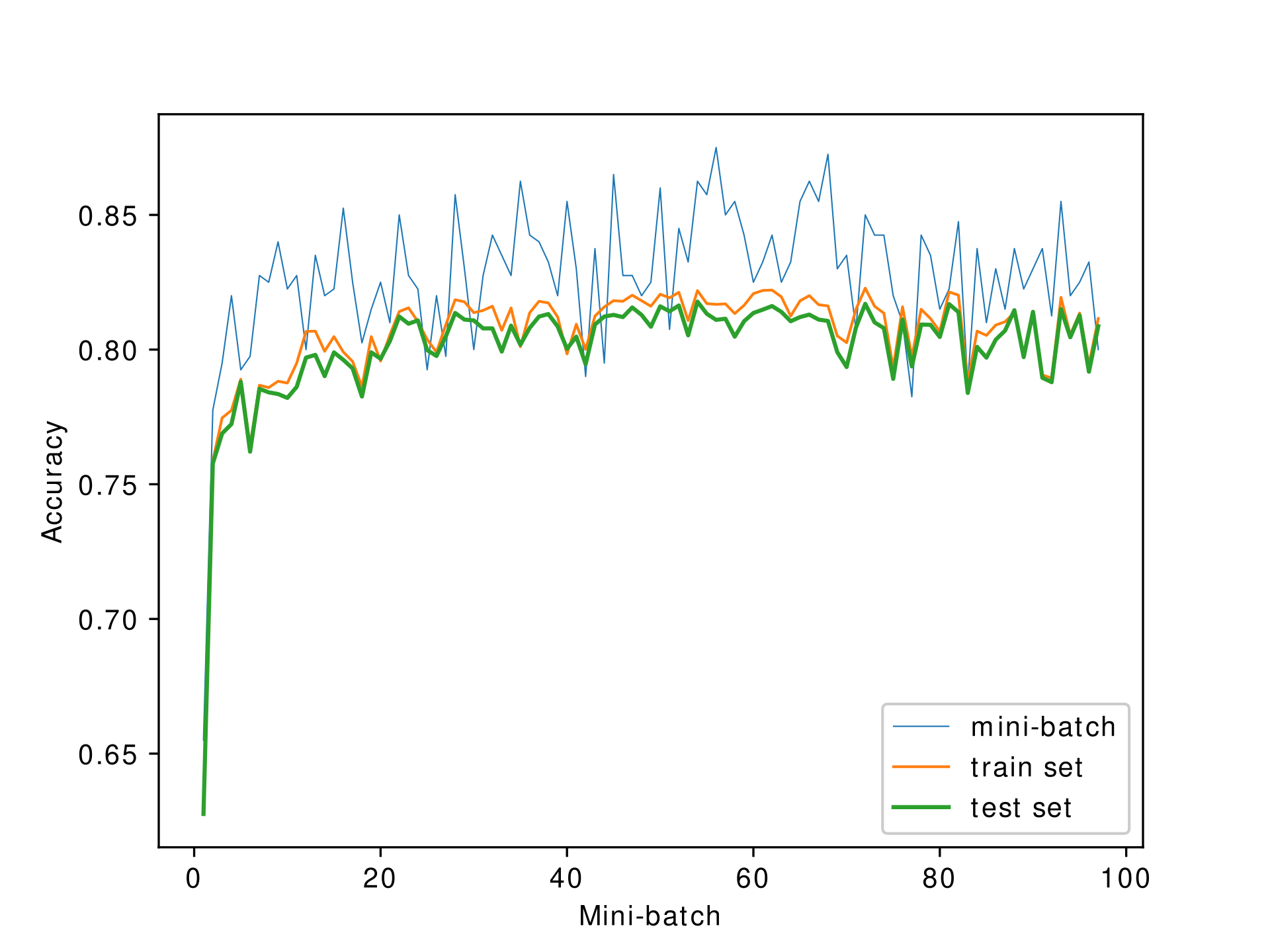}
  \captionof{figure}{Accuracy over number of mini-batches.}
  \label{fig1}
\end{minipage}
\begin{minipage}{0.37\linewidth}
  \centering
  \caption{Accuracies Network vs \ripper.}
  \label{tab1}
  \medskip
  \begin{tabular}{lrr} 
  \hline
  Dataset       & Network & RIPPER  \\ 
  \hline
  adult         & 0.7645  & 0.8234  \\
  airline       & 0.7783  & 0.7820  \\
  alpha-bank    & 0.8735  & 0.8734  \\
  bank          & 0.8830  & 0.8924  \\
  breast-cancer & 0.6946  & 0.6958  \\
  credit-g      & 0.7020  & 0.7110  \\
  hepatitis     & 0.7925  & 0.8129  \\
  kr-vs-kp      & 0.8477  & 0.9919  \\
  mushroom      & 0.9745  & 1.0000  \\
  vote          & 0.9290  & 0.9540  \\
  \hline
  \end{tabular}
\end{minipage}
\end{table}
\vspace{-0.3em}

\ripper performs better than the rule network on nine out of ten datasets and almost the same on the remaining one. For half of the datasets, the rule network performs at least 2\% worse and for two the difference even exceeds 5\%. In both the adult dataset and the kr-vs-kp dataset (king rook vs king pawn in chess endgame) the rule network seems to suffer from an unsuitable initialization of the rules. 

The training time of the rule network is ten times longer than for \ripper. Note, however, that a good level of accuracy is typically reached much earlier (e.g., at about 25\% for the \textsf{adult} dataset in Fig.~\ref{fig1}). Finding a good stopping criterion for the iterative refinement is a question we are currently working on.

\section{Conclusions}
We have proposed a technique to use mini-batches for learning rule sets. Even though the approach is still outperformed by \ripper, we think that the achieved performance level justifies a deeper investigation.
In future work, we thus plan to improve the mini-batch approach by the usage of gradient descent or another more targeted search of flips and the establishment of an early-stopping criterion.

\bibliographystyle{splncs04}
\bibliography{bibliography}

\begin{thebibliography}{1}
\providecommand{\url}[1]{\texttt{#1}}
\providecommand{\urlprefix}{URL }
\providecommand{\doi}[1]{https://doi.org/#1}

\bibitem{DBLP:conf/icml/Cohen95}
Cohen, W.W.: Fast effective rule induction. In: Prieditis, A., Russell, S.J.
  (eds.) Machine Learning, Proceedings of the Twelfth International Conference
  on Machine Learning, Tahoe City, California, USA, July 9-12, 1995. pp.
  115--123. Morgan Kaufmann (1995). \doi{10.1016/b978-1-55860-377-6.50023-2}

\bibitem{DBLP:journals/datamine/DembczynskiKS10}
Dembczynski, K., Kotlowski, W., Slowinski, R.: {ENDER:} a statistical framework
  for boosting decision rules. Data Min. Knowl. Discov.  \textbf{21}(1),
  52--90 (2010). \doi{10.1007/s10618-010-0177-7}

\bibitem{DBLP:series/cogtech/FurnkranzGL12}
F{\"{u}}rnkranz, J., Gamberger, D., Lavrac, N.: Foundations of Rule Learning.
  Cognitive Technologies, Springer (2012). \doi{10.1007/978-3-540-75197-7}

\bibitem{DBLP:journals/datamine/KosinaG15}
Kosina, P., Gama, J.: Very fast decision rules for classification in data
  streams. Data Min. Knowl. Discov.  \textbf{29}(1),  168--202 (2015).
  \doi{10.1007/s10618-013-0340-z}

\bibitem{mr:ECML-PKDD-2020}
Rapp, M., Loza~Mencía, E., F\"urnkranz, J., Nguyen, V.L., H\"ullermeier, E.:
  Learning gradient boosted multi-label classification rules. In: Proceedings
  of the European Conference on Machine Learning and Knowledge Discovery in
  Databases (ECML/PKDD). Springer-Verlag (2020),
  \url{http://arxiv.org/abs/2006.13346}

\bibitem{DBLP:conf/icml/RuckertK03}
R{\"{u}}ckert, U., Kramer, S.: Stochastic local search in k-term {DNF}
  learning. In: Fawcett, T., Mishra, N. (eds.) Machine Learning, Proceedings of
  the Twentieth International Conference {(ICML} 2003), August 21-24, 2003,
  Washington, DC, {USA}. pp. 648--655. {AAAI} Press (2003),
  \url{http://www.aaai.org/Library/ICML/2003/icml03-085.php}

\bibitem{DBLP:journals/corr/YangYC17}
Yang, F., Yang, Z., Cohen, W.W.: Differentiable learning of logical rules for
  knowledge base completion. CoRR  \textbf{abs/1702.08367} (2017),
  \url{http://arxiv.org/abs/1702.08367}

\end{thebibliography}

\begin{appendices}
\renewcommand{\thesection}{\appendixname~\Alph{section}}
\section{Network Structure}
  \vspace{-0.6em}
  \includegraphics[width=\textwidth]{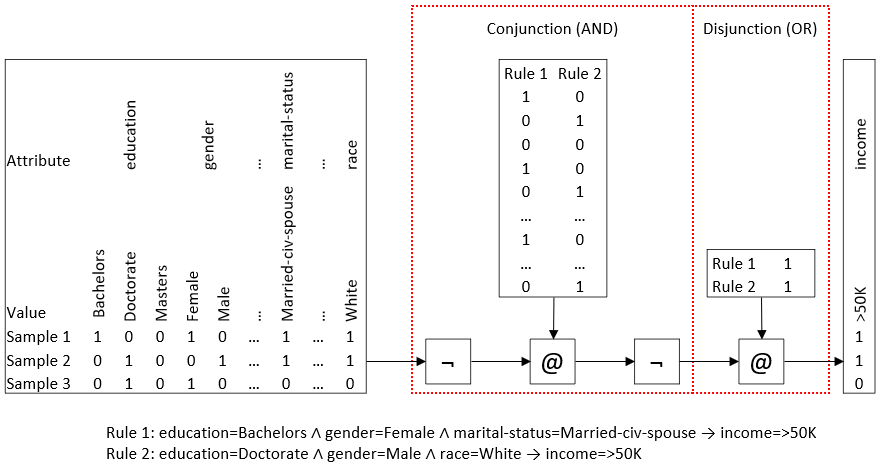}
  \captionof{figure}{Processing of the data through the network. The network is shown within the red dashed line, split into \texttt{AND} and \texttt{OR} layer. $\lnot$ represents inversion and $@$ boolean matrix multiplication. Two example rules learned on the \textsf{adult} dataset are provided. Note that for better clarity, \texttt{False} and \texttt{True} have been replaced by 0 and 1 respectively.}
  \label{fig2}
  \centering
\end{appendices}

\end{document}